\documentclass[12pt]{spieman}
\usepackage{amsmath,amsfonts,amssymb}
\usepackage{graphicx}
\usepackage{verbatim}
\usepackage{setspace}
\usepackage{tocloft}
\usepackage{longtable}
\usepackage{comment}
\usepackage{booktabs}
\usepackage[numbers]{natbib}
\usepackage{notoccite}
\usepackage{algorithm}
\usepackage[noend]{algpseudocode}
\usepackage{lineno}

\title{Bringing the Algorithms to the Data - Secure Distributed Medical Analytics using the Personal Health Train (PHT-meDIC)} 

\author[e*]{Marius de Arruda Botelho Herr}
\author[a]{Michael Graf}
\author[a]{Peter Placzek}
\author[f]{Florian König}
\author[e]{Felix Bötte}
\author[e]{Tyra Stickel}
\author[a]{David Hieber}
\author[e]{Lukas Zimmermann}
\author[e]{Michael Slupina}
\author[e]{Christopher Mohr}
\author[e]{Stephanie Biergans}
\author[b,e,f]{Mete Akgün}
\author[b,c]{Nico Pfeifer}
\author[a,b,d,f]{Oliver Kohlbacher}

\affil[a]{Institute for Translational Bioinformatics, University Hospital Tübingen, Tübingen, Germany}
\affil[b]{Institute for Bioinformatics and Medical Informatics, University of Tübingen, Tübingen, Germany}
\affil[c]{Methods in Medical Informatics, Department of Computer Science, University of Tübingen, Germany}
\affil[d]{Applied Bioinformatics, Department of Computer Science, University of Tübingen, Germany}
\affil[e]{Medical Data Integration Center, University Hospital Tübingen, Tübingen, Germany}
\affil[f]{Medical Data Privacy and Privacy-Preserving ML on Healthcare Data, Department of Computer Science, University of Tübingen, Germany}

\cftpagenumbersoff{figure}
\cftpagenumbersoff{table} 
\begin{document} 
\maketitle

\begin{abstract}
The need for data privacy and security -- enforced through increasingly strict
data protection regulations -- renders the use of healthcare data for machine learning
difficult. In particular, the transfer of data between different hospitals is often not permissible
and thus cross-site pooling of data not an option. The Personal Health Train (PHT) 
paradigm proposed within the GO-FAIR initiative implements an 'algorithm to the data'
paradigm that ensures that distributed data can be accessed for analysis without transferring any sensitive data. 
We present PHT-meDIC, a productively deployed open-source implementation of the PHT concept. Containerization allows us to easily deploy even complex data analysis pipelines
(e.g, genomics, image analysis) across multiple sites in a secure and scalable manner.
We discuss the underlying technological concepts, security models, and governance processes.
The implementation has been successfully applied to distributed analyses of large-scale data, including applications of deep neural networks to medical image data.
\end{abstract} 


{\noindent \footnotesize\textbf{*}Marius de Arruda Botelho Herr,  \linkable{marius.de-arruda-botelho-herr@uni-tuebingen.de} }

\begin{spacing}{1}   


\section{Introduction}
\label{sect:intro}  
During the last decades, the advantages of artificial intelligence (AI) became evident in many applications that affect our daily lives (e.g., smart assistants, chatbots, application processes) \cite{Dignum2018}.
The rapid progress in these fields proved especially beneficial for big data analysis in healthcare \cite{bigdataHC}. Machine learning models that support clinical practice \cite{ai-healthcare} in the form of treatment decision support systems (TDSS) are just one prominent example, paving the way towards a 'learning healthcare system' \cite{Overhage2007,Belard2016}.

The implementation of learning systems in healthcare environments poses several challenges. Successful machine learning typically relies on large amounts of well-curated data, available in a consistent format - and located in a single place. Additionally data protection concerns, semantic heterogeneity, and data governance issues need to be addressed to make the analysis of big healthcare data a matter of routine. The naive way to address these issues is the integration of disparate data silos \cite{data-silos} into central data repositories.  Examples for this are databases for genomic analysis, such as the 100000 Genomes Project \cite{Torjesenf6690} and the  All of Us Research Program \cite{doi:10.1056/NEJMsr1809937}. Even though they are a convenient solution, a concern with central data repositories is the protection of the patient data, in particular in Europe under the GDPR \cite{EUdataregulations2018}. Central databases pose a greater risk of re-identifying patient data, due to their centralized nature and the possibility to combine the information with other publicly available data sets \cite{cloud_breach, hippa_fail}.
While applications for the distributed analysis of healthcare data have been established and applied successfully in many instances \cite{SHRINE, OHDSI, datashield, medco, sharemind, i2b2} and successfully implemented in many studies, there are particular challenges arising if structured clinical data is augmented by high-volume data arising from (raw) genomic data, imaging, or biosignal data. Analyzing this data usually requires complex analysis pipelines that cannot be readily integrated into the federated query concepts of the approaches mentioned above.

As part of the German national Medical Informatics Initiative (MII) \cite{MII,difuture}, German university medical centers have chosen a distributed approach to data management and governance. Since 2018 data integration centers (DIC) are being established, with the task of providing secure and reliable access to pseudonymized healthcare data for research, while maintaining control over the data by local authorities. As part of this infrastructure distributed analysis tools will be used in order to analyse healthcare data in a decentralized fashion, and overcome the issues outlined above.

The Personal Health Train (PHT) is a framework for distributed analysis proposed by the GO-FAIR initiative \cite{GoFair,Fair}. The basic concept of the PHT is that the analysis algorithm (wrapped in a 'train') travels between multiple sites (so-called 'train stations'), which are securely hosting the data. These trains are implemented as light-weight containers. Container-based virtualization enables the rapid deployment of complex software pipelines (e.g., for genomics or image analysis) without requiring any additional software installation.  These concepts have been described in the literature before \cite{PHT_1, PHT_2, PHT_3, PHT_LZ_1}, but production-ready implementations have been lacking so far.
A recently published federated PHT architecture PADME \cite{Welten_PADME_1} provides more details regarding the application but details regarding the security concept are lacking.
Another federated PHT vantage6 \cite{moncada-torres2020vantage6} is mainly focusing on image analysis and details on data privacy are similar to PADME.

In this publication we present our software architecture for the PHT with a special focus on security features as well as an open-source implementation (PHT-meDIC) which is in productive use at German university hospitals. We demonstrate with our tool PHT-meDIC how modern cloud techniques can be leveraged for complex distributed, privacy-preserving medical data analytics. Specifically, we make the following contributions:
\begin{itemize}
\item an architecture and security framework for the PHT
\item a semantically integrated data access mechanism based on HL7/FHIR
\item implementation of data governance processes within this framework
\item two showcases demonstrating distributed machine learning on big data with the PHT
\item a freely available open-source implementation of these features
\end{itemize}

Our proposed architecture is split into central and local components. To minimize local administration, our framework uses container technologies to enable easy distribution of arbitrarily complex analysis pipelines.
Compared to other proposed PHT solutions, our solution implements security-by-design. Our security protocol guarantees that the code executed at each train station can not be manipulated in transit or during any other stage of the process. Additionally the results are encrypted in transit to protect them from interception by anybody but participants conducting a specific study. Decryption of the result is only possible for the creator of the algorithm and participating study sites.
Exemplifying secure privacy preserving count queries, shows the possibility of how our PHT-meDIC architecture can simply be extended with additional data privacy preserving methods.

To the best of our knowledge this represents the first distributed decentral PHT implementation ensuring security, data privacy and all required governance documents to deploy the platform productively.

\section{Methods}

\subsection{PHT-meDIC Concept}
The PHT-meDIC implementation enables the secure analysis of distributed bio-medical data. 
Such analyses can consist of a wide range of complex bioinformatics pipelines, training machine learning models such as deep neural networks, or secure count queries and summary statistics.
An analysis in the context of the PHT-meDIC is defined as arbitrary Python or R code submitted by researchers to be containerized and executed sequentially in distributed train stations. The sensitive input data always remains under controlled access by each participating site (Figure \ref{fig_short}).

The main benefit of the PHT architecture compared to other federated analysis systems is its ability to transport rather complex pipelines consisting of many different tools to the sites without local software installation. PHT achieves this through the use of lightweight container-based virtualization. By now, the single point of communication between stations and PHT central services is reached utilizing docker pull and push commands. The security concept of the PHT-meDIC aims to minimize risks while still allowing the full functionality and flexibility of our PHT-meDIC architecture.

\textbf{\begin{figure}[h!]
	\centering
    \includegraphics[width=\textwidth]{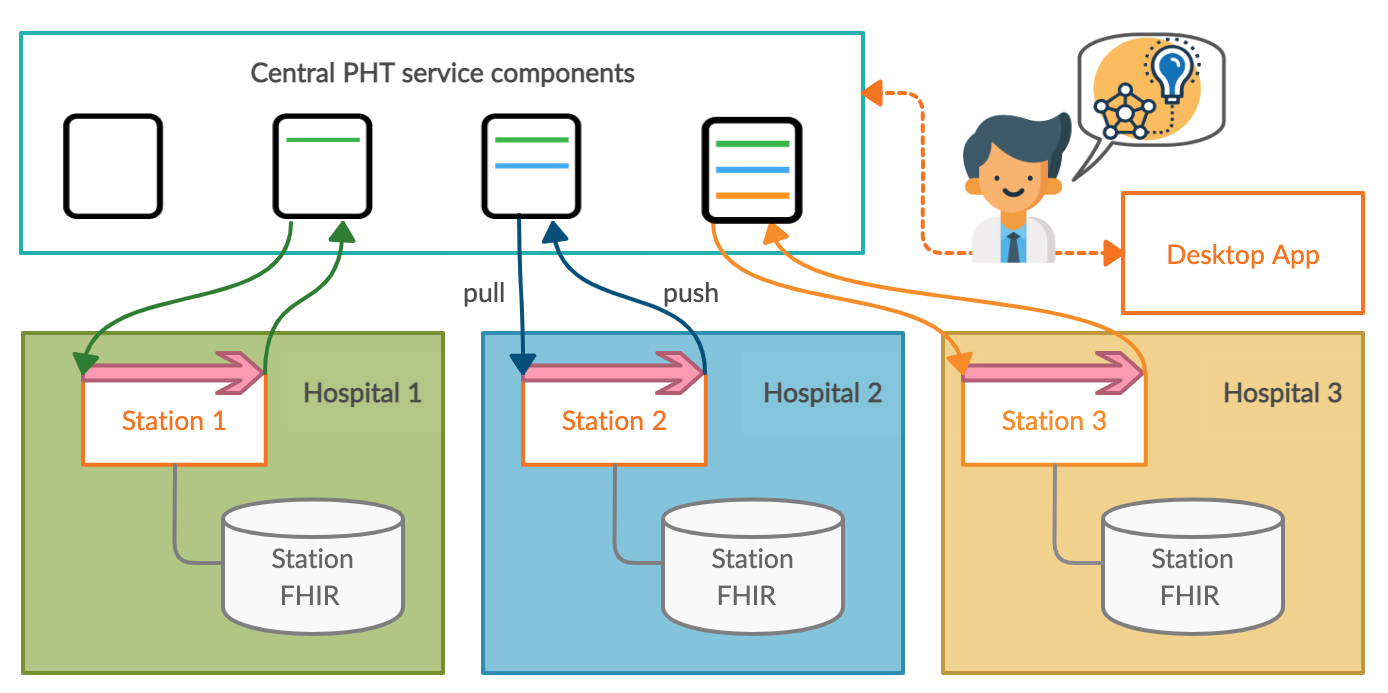}
	\caption{Simplified overview of execution and update of results at three stations. In total seven central components are included as central services.
	Big pink arrows indicate the execution of trains locally at each hospital. Lines within the boxes indicate the update of results. Colored arrows between registry and hospitals illustrate the push and pull commands.}
	\label{fig_short}
\end{figure}}

\subsection{Related work} 
Solutions in distributed learning settings such as euroCAT \cite{euroCAT}, ukCAT \cite{ukCAT}, and bwHealthApp \cite{BWHealth} overcome the problem of data sharing while keeping the data local. These are typically developed for specific use cases and are not generic enough to be easily extended for different kinds of analyses or programming languages. SHRINE/i2b2 is also an open-source health data warehouse, used in various worldwide projects \cite{SHRINE}. However, it is limited on the execution of queries and therefore mostly used for cohort selection for clinical trials. Another open-source and widely established project, OMOP/OHDSI \cite{OHDSI} relies on the common Common Data Model (CDM) and therefore limits the input data to be processed.
DataSHIELD \cite{datashield}, another open-source distributed learning framework, is more generic and extensible. However, DataSHIELD is limited to the R programming language and only functions installed by each data provider can be executed, making it difficult to deploy the complex pipelines often required for genomics or imaging.

All published PHT implementations~\cite{moncada-torres2020vantage6, Welten_PADME_1} are developed based on regulations and individual research needs, but are all following the PHT GO:FAIR manifesto as guidance.
Our PHT-meDIC has clearance to operate in productive use at German University hospitals, and our security concept was presented and shared in 2019.
Compared to PADME, the local decryption of results with the desktop app and our central TR service add additional security.
Our developed user interface allows us to extend it for easy usage for varying disciplines making it available to not only computer scientists.
Vantage6 utilizes R and Python clients, which require a technical understanding to use.

Secure multi-party computation frameworks (SMPC) provide excellent security and privacy for federated analyses. They operate under malicious threat models and compute global results on fully-encrypted input data.
Typically, SMPC solutions are domain-specific, highly specialized on particular tasks and are limited in their analysis functionalities by design. In terms of scalability most solutions still have some limitations due to significant performance overheads in securing the analysis.
Sharemind MPC \cite{sharemind} is proprietary software which is limited to statistics in R and not trivially extendable.
The latest SMPC frameworks have overcome scalability limitations and can even operate fast on big data \cite{mete_smpc,Drynx}. MedCO \cite{medco} uses a combination of several protocols and techniques \cite{Drynx, UnLynx} to allow secure analysis on distributed data but is currently limited to tranSMART \cite{transmart} as primary data storage. A newly released framework, FAHME \cite{lausanne3} from the same researchers as MedCO is based on lattigo \cite{lattigo}, but limited regarding analysis functionalities due to fully homomorphic encryption.
 
\subsection{PHT-meDIC architecture}
\begin{figure}[h!]
	\centering
    \includegraphics[width=\textwidth]{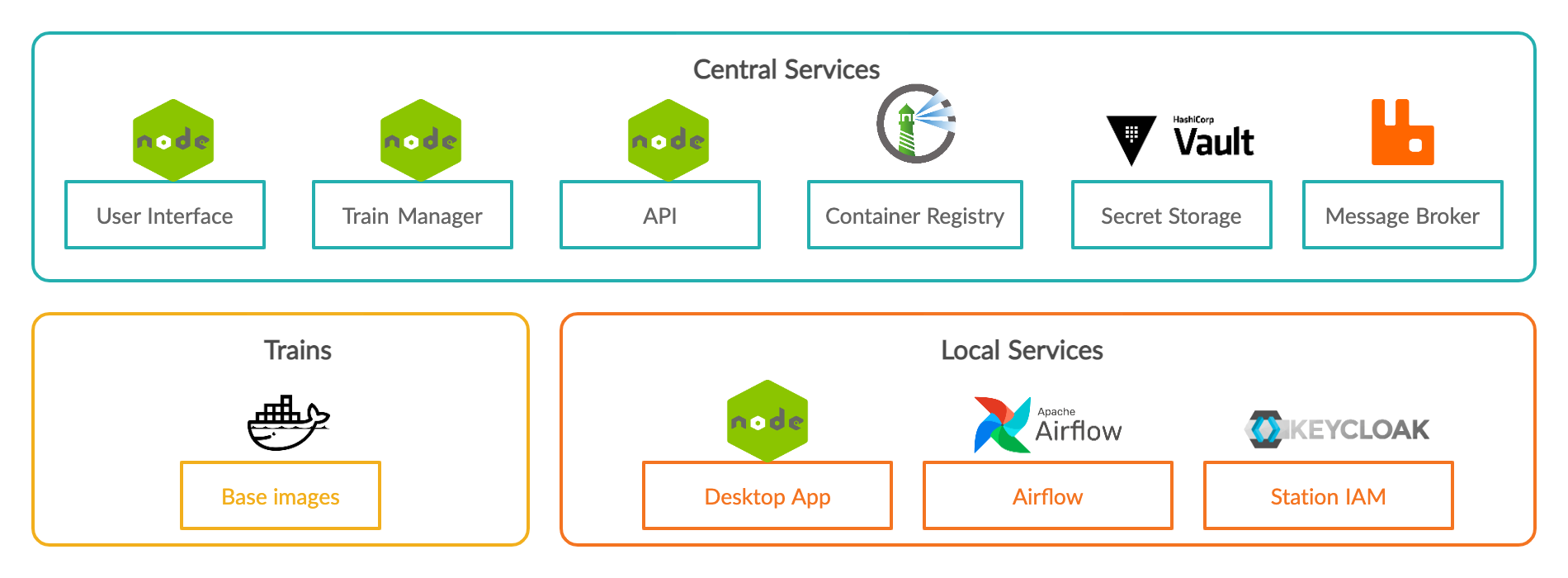}
	\caption{Existing open-source service components of the PHT-meDIC are displayed with the icon of the service. Self-developed service components for the PHT-meDIC contain icons of the programming language in which we developed those.} 
	\label{fig_services}
\end{figure}

The architecture of the PHT-meDIC consists of central and local services that enable iterative execution of analysis code, which is wrapped in trains, as illustrated in Figure \ref{fig_services}.
The \textit{User Interface} (UI) is the central entrypoint into our platform. Users can submit study requests (project proposal), analysis code for these proposals and manage participation in analyses. 
The \textit{Desktop App} is an application installed on a users PC that can be used to generate private and public key pairs. These key pairs can then be used to cryptographically sign the contents of a train.
The \textit{Train Manager} (TM) summarizes several tasks: Train Building (TB), Train Routing (TR) and Result Extraction (RE). TB is the process of creating valid 'trains' based on the uploaded files and creates the required configuration file to execute a 'train'. TR is the task to interact with the Secret Storage and the \textit{Container Registry} (CR) to make the submitted train available for each station. After successful execution, the RE makes the encrypted results available to download within the UI.
The \textit{API} is a REST interface to manage central resources and trigger processes between services.
The container registry (CR), is used for storage and distribution of 'train' (see \ref{Train}) containers. We use the open-source registry Harbor \cite{harbor} to provide this functionality. For storage of sensitive data such as train routes or submitted public keys of stations and users we utilize the open-source tool Vault \cite{vault}. Internal communication between central services is based on the asynchronous open-source message broker RabbitMQ \cite{rabbitmq}.
Trains are based on base images, which include all study specific software dependencies and the train library. The train library includes the security protocol and domain logic of trains.
The station is a local service, based on Airflow \cite{airflow}, hosted locally at each study site providing access to the local data in a secure environment and enabling the execution of trains under the security protocol. A station only requires communication to the CR in order to participate in the analysis. 
Encrypted results can be downloaded from the UI and decrypted using the \textit{Desktop App} with the users private key to obtain results locally.

\subsubsection*{Train components}\label{Train}
A train is based on Docker \cite{docker} container technologies and is self-contained with all required software dependencies to run the analysis code. This enables local execution without any additional installation efforts.
We provide a set of trusted base images in a publicly accessible repository containing all packages and libraries to execute the submitted algorithm. 
If the existing base images do not fulfill the user's requirements, users can propose modifications to existing images or create new ones and request this new image to be approved as a base image. All base images are regularly scanned for software vulnerabilities.
The user submitted algorithm and FHIR search query are stacked on top of the base image.
These files are static and do not change during execution. Results of the train are encrypted at rest and updated iteratively during execution. A configuration file containing the public keys of stations and users, the hash of static files, result signatures of previous executions and the encrypted symmetric key is included in the train image. This file is used perform the security protocol.

\subsubsection*{Train submission and execution}
After acceptance of a study proposal by each station, the user uploads an algorithm and specifies the participating stations. 
During the submission process, the user is required to sign the hash of the uploaded algorithm using the Desktop App. After the train configuration is finished a building command is sent to the message broker. TM consumes this building command, after requesting the algorithm from the API and obtaining all required public keys from Vault the train image is built and the route stored in Vault. The newly built train is pushed to the incoming train repository of the CR, only accessible by the TR. When the users issues the start command in the UI, the train is moved to the initial station repository based on the route stored in vault and is ready for executions. Station repositories can only be accessed by the respective station (and the TR for distribution). If a train is available, the station administrator \textit{pulls} and executes the train using airflow. Required resources to run the analysis are specified in the station user interface.

After successful execution, the station \textit{rebases} and \textit{pushes} the train back to the stations private repository in the CR. The TR process moves the updated train to the next station repository. A train never stores any of the data provided for analysis - only encrypted results are stored and distributed. Every station on the route follows the same procedure. When the train has reached the end of it's route, it is moved to the outgoing repository. The RE scans the project for successfully executed trains, extracts encrypted results, and transmits these to the UI. The UI notifies the user, who can download the results and decrypt them locally using the Desktop App.

\subsubsection*{Train stations}\label{stat}
Stations are operating at each hospital and are based on Apache Airflow \cite{airflow}. This open-source workflow management platform allows persistent and monitored execution of incoming trains as tasks defined as directed acyclic graphs (DAGs). The only required communication channel between stations and central services is the CR.
Within the Airflow web interface, a station administrator can specify which train should be executed and provide addtional execution configurations. The station administrator can set configurations (e.g. data sources, station identifiers, or keys) globally for all incoming trains or individually for each train. Any clinics study system (e.g. FHIR, RedCAP, or tranSMART) are independent operated and can be used as data sources. After pulling a train from the CR the station first verifies that no static files were manipulated during transit. If there were previous executions of the train, the previous results and execution order is validated using the hash chain based digital signature. After successfull verfication of the train the encrypted previous results are decrypted and the algorithm executed using the provided data. The updated results are encrypted and stored in the outgoing train image. To overcome the limitation of 127 layers of a docker image \cite{docker_limit} a rebasing is required to maintain a maximum of $n + 1$ layers in the resulting image.

\subsection{Security and privacy of trains}
The highly sensitve nature of medical data, i.e. mental health or genetic data, demands the highest effort in keeping this information private. The security protocol (supplementary material \ref{supp:security}) of the PHT-meDIC aims to provide security and privacy without restricting functionality. We consider the following threat models for our services: 
\textit{Container registry} and the \textit{user} can delete, modify or change parts of the train to obtain sensitive information. Therefore they are considered malicious. The \textit{stations} are considered to be honest-but-curious. They can attempt to infer sensitive information from results stored in the train. The \textit{secret storage} Vault and the \textit{Train Manager} (train building, train routing and result extraction) are assumed to be trusted parties.
Several security features ensure only registered users have access and only reviewed and approved trains can be executed. Furthermore, we ensure that the submitted code from a user remains unchanged during the whole execution and verify all previous executions of the train. A combination of symmetric and asymmetric encryption, known as envelope encryption, allows for accelerated encryption and decryption of results. As an additional security measure trains do not have network access and never directly access data from local sources. Read only access to the input data is provided to the train by the station.

A honest-but-curious station can access results of previously visited stations and the route. A station is capable of performing model inversion attacks \cite{modelinversion} or membership inference attacks \cite{membershipinference}. However, these kinds of attacks are rarely successful on models with a high-dimensional input space \cite{mlattacks}. Calculating simple count queries over patient data, may provide an attacker enough information to re-identify patients or hospitals \cite{Inference_attack, count_queries}. To solve this the train library is extended to support paillier homomorphic encryption \cite{pht-sec-add}. Other privacy preserving methods can be applied. Stations can also compute statistics based on the configuration file, such as the number of times a specific station was involved in the previous analysis. However, since only pseudo identifiers are present in the train it is unlikely that they can be matched to actual identifiers. 
The main reason stations are also honest-but-curious is that they have to use real input data during train execution. Apart from this requirement, stations can be assumed to be malicious. They can modify the train such as changing the algorithm for malicious purposes. However, such modifications will be detected by the next station preventing any modified trains from being executed. Without having access to the user's private key no station can modify the train content or signature without being detected by the pre-run protocol (supplementary material table \ref{train_pre_run}).

A malicious container registry (CR) can access all station repositories and images, and CR can move an image from one repository to another. Stations detect this change by verifying the route using the hash chain based digital signature created by each previous stations.  Stations detect any modification on the algorithm, the route, and the query because it is impossible to generate a valid user signature for malicious parties. Any changes to encrypted partial results done by CR are also detected because stations digitally sign their intermediate results using their private key.
Furthermore, it is not possible to change the image for CR, such as adding malicious software. The base image is taken from the list of trusted images which is prepared by governance. Thus it is simple to detect any changes in the image using docker technology. 

Even in the case of collusion of the user, CR, and stations, it is not possible to change the algorithm, query, route, and docker image. Because the algorithm, query, and route are signed by the trusted TM and the list of trusted images is stored in TM.

\section{Demonstration showcases}
The following two example trains are proof-of-concepts of the functionality of our platform. Furthermore they demonstrate how two unrelated tasks on different high-volume medical input data can be solved using the PHT-meDIC. The image analysis compares centralized with decentralized results. Within the bioinformatics pipeline the outcomes are further processed with privacy preserving methods to prevent later stations from examining previous results.

\subsection{Bioinformatics pipeline}\label{exp:nf-core}
\subsubsection*{Problem and input}
The Human Leukocyte Antigen (HLA) system is a gene complex that encodes the major histocompatibility complex (MHC) proteins in humans. MHC genes are highly polymorphic and have many variations across humans \cite{hla}. It is pleiotropic, and its main functionality is to support T cells to recognize foreign proteins (antigens), an essential process in adaptive immunity. Therefore it is a significant driver for the initiation and coordination of immune responses \cite{hla_hiv}. Determining HLA types of subjects is important in several areas: therapeutics can be affected by HLA type, potential transplants require a match of donor and recipient HLAs, and certain infectious diseases can progress more rapidly \cite{hla_hiv}.

Traditionally HLA typing is done by medical centers. Results are usually returned within medical reports and are per default not accessible for further analysis. Approaches like the German Human Genotyping and Phenotyping Archive (GHGA \cite{GHGA}) are currently aiming to solve this problem. If clinical sites have genome sequencing data of patients available, it is usually not trivial to analyze the raw data ad hoc. By developing standardized and reproducible bioinformatics analysis pipelines, the nf-core \cite{nf-core} initiative aims to overcome this bottleneck. Usually, these pipelines are run by medical centers and not the hospital itself. Using the PHT-meDIC, it is now possible to run these complex analyses and combine the results across sites.

\subsubsection*{Analysis}
We used 216 next-generation sequencing (NGS) samples (supplementary material table \ref{sup:subjects}) consisting of Illumina HiSeq 2000 and Genome Analyzer II exome sequencing runs from the 1000 Genomes Project \cite{RN8} to demonstrate the use of such pipelines. We executed an nf-core HLA typing pipeline \cite{hla-code} over three stations with the PHT-meDIC.
Subject data is mapped to FHIR profiles, containing references (URIs) to locally stored NGS data, which can be processed by the train.
The base image of the train contains \textit{nf-core/hlatyping} \cite{hla-code} software dependencies and the Python algorithm requirements. Following PHT principles the algorithm and query are stacked on top of the base image. The algorithm has two tasks:
\begin{enumerate}
\item Compute HLA types of subjects and securely determine the number of patients with class I HLA type B*35:01 (fast progression of HIV/Aids \cite{hla_hiv}).
\item Plot the absolute most frequent 15 HLA types of all provided samples.
\end{enumerate}

\subsubsection*{Results}
\textbf{Task 1:} Locally decrypting the results reveals the homomorphically encrypted result. Decrypteing this number with the paillier private key reveals $24$ occurrences of Class I allele \textbf{B*35:01} over all stations.

\begin{figure}[h!]
  \centering
    \includegraphics[width=.5\textwidth]{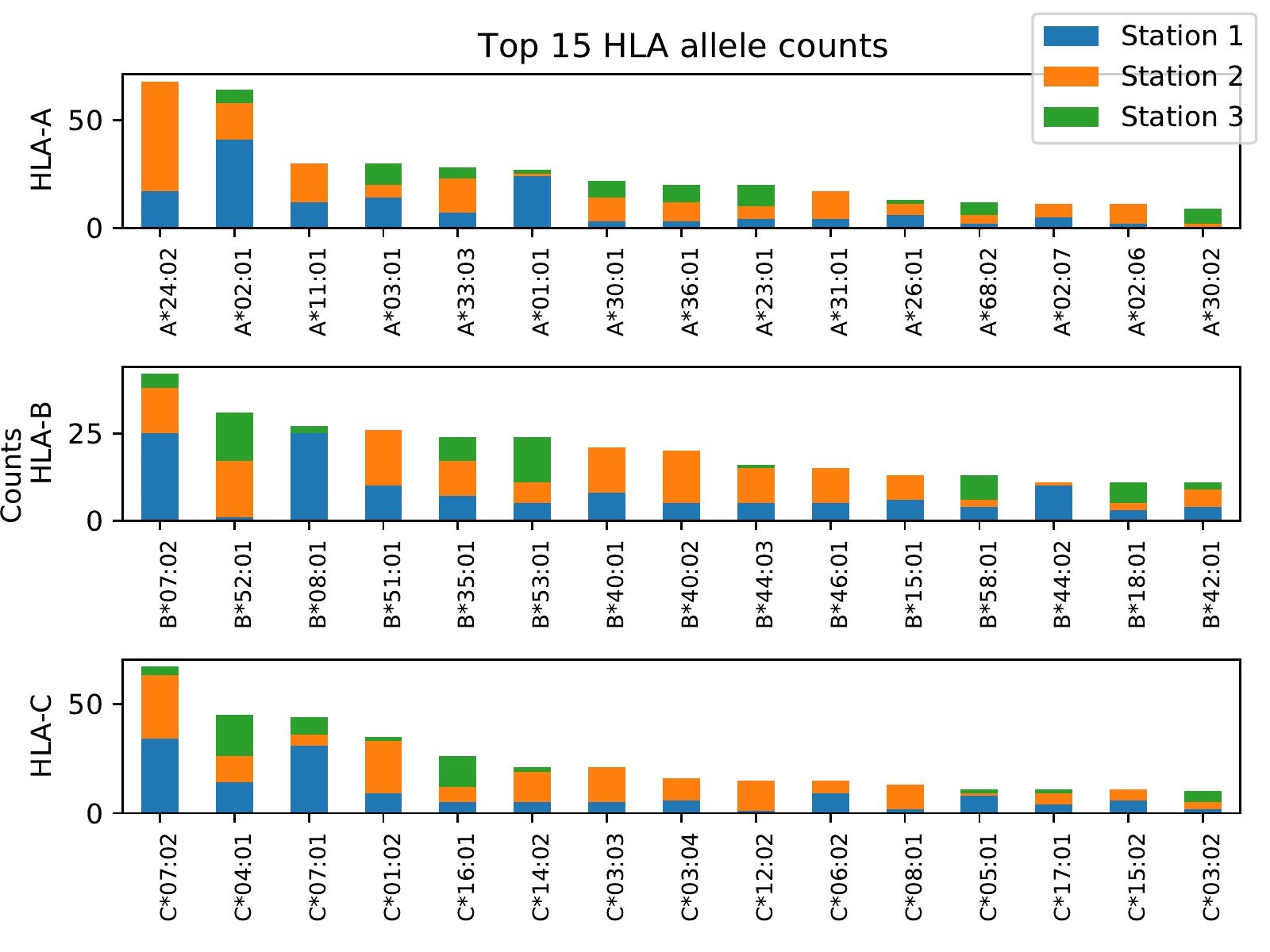}
	\caption{Results of the second task of the bioinformatics pipeline train. This plot shows three different HLA classes and their allele counts and ordered by number of occurrences. Colors indicate data originating from different stations.}\label{exp:train1_taskb}
\end{figure}

\noindent \textbf{Task 2:} The resulting figure of the second task (plotting the most frequent HLA types) is displayed in Figure \ref{exp:train1_taskb}. The plot contains the most 15 common HLA types across the stations included in the analysis.

\begin{table}[h!]
\centering
\begin{tabular}{|l|l|l|}
\hline
\textbf{Station} & \textbf{\# of samples}  & \textbf{Execution time} \\ \hline
1      & 84  & 13h 46min \\ \hline
2     & 94    & 24h 26min \\ \hline
3 & 38 & 6h 37min \\ \hline
Total & 216 & 44h 49min\\ \hline
\end{tabular}
\label{train1:time}
\caption{Station-wise and total number of samples and execution time of the nf-core HLA typing train}
\end{table}

\noindent
The total execution time of the HLA tytping demonstration train was $44$h and $49$ min on a Ubuntu $20.04.1$ LTS cloud instance (table\ref{train1:time}) with $28$ Intel(R) Xeon(R) Gold 6140 CPU @ $2.30$ GHz cores and $256$ GB memory. The final results have a size of $22$ KB and took below $1$ s at each station to decrypt and encrypt using the PHT-meDIC security protocol.

\subsection{Image analysis}
\subsubsection*{Problem and input}
The  International Skin Imaging Collaboration (ISIC) 2019 is a dermoscopic image analysis benchmark challenge containing more than 25 thousand images of eight different skin cancer types \cite{isic1,isic2,isic3}. The goal of the challenge is to support research and development of algorithms for automated melanoma diagnosis by providing data with an annotated ground truth by medical experts. For demonstration purposes, we use the publicly available algorithm of the challenge champions \cite{GESSERT2020100864}.
In this showcase we demonstrate the feasablity of performing image anaylsis with the PHT-Medic. Additionally we compare the performance of the model trained in a distributed setting with the traditionally trained model.

\subsubsection*{Analysis}
The task of the analysis is to classify eight different types of skin cancer. We use all publicly available training images and train an efficientnet-b6 Deep Neural Network (DNN). In the centrallized analysis setting, we use the provided 5-Fold Cross-Validation (CV) splits from the challenge's first place submission \cite{GESSERT2020100864}. In the distributed setting, we perform a 5-Fold CV at each station locally using the identical fold splitting as within the central analysis setup. The distribution of input data over each station is shown in table \ref{exp:train2}. The algorithm takes class imbalance into account by assigning weights to each class according to its availability. Input data is mapped to FHIR and processed by the train as in the first experiment.

\begin{table}[h!]
\centering
\begin{tabular}{|l|llllllll|l|}
\toprule
\textbf{Site}      & \textbf{MEL}  & \textbf{NV}    & \textbf{BCC}  & \textbf{AK}  & \textbf{BKL}  & \textbf{DF}  & \textbf{VASC} & \textbf{SCC} & \textbf{Total} \\ \midrule
\textbf{Station 1} & 1507 & 4291  & 1107 & 289 & 874 & 79 & 84   & 209 & \textbf{8440}  \\ \midrule
\textbf{Station 2} & 1507 & 4291  & 1107 & 289 & 874 & 79 & 84   & 209 & \textbf{8440}  \\ \midrule
\textbf{Station 3} & 1508 & 4293  & 1109 & 289 & 876 & 81 & 85   & 210 & \textbf{8451}  \\ \midrule
\textbf{Central}     & \textbf{4522} & \textbf{12875} & \textbf{3323} & \textbf{867} & \textbf{2624} & \textbf{239} & \textbf{253}  & \textbf{628} & \textbf{25331} \\ \bottomrule
\end{tabular}
\caption{Distribution of melanoma imaging data over three stations and total number of samples of each different class of skin lesions. The cancer class abbreviations can be found in the supplementary material \ref{supp:notations}.}
\label{exp:train2}
\end{table}

\subsubsection*{Imaging train results}
The efficientnet-b6 model, trained with all data at one station has a final averaged accuracy of $0.682$ over 5-folds. Accuracy is continuously calculated class-wise and averaged with all classes. The distributed model, trained with only 20 epochs at each station for comparison, has a final averaged accuracy of $0.681$. Even with equal distribution of each class at each station locally, the drop in the performance of the model at station three can be attributed to crucial forgetting (or \textit{Catastrophic Forgetting})~\cite{goodfellow2013empirical,kemker2018measuring,mccloskey1989catastrophic}, which can depend on data quality differences or data availability. 

The execution time of the second example train on a cloud Ubuntu $20.04.1$ LTS instance with $2$ \textit{Nvidia TeslaV100} GPUs, $16$ Intel(R) Xeon(R) Gold 6140 vCPU @$2.30$ GHz cores and $170$ GB memory is $37$ h $9$ m and $8$ s in total. The encryption and decryption of results over three stations took in total $5$ m and $31$ s. Station wise execution time and size can be seen in Table \ref{exp2:time}.

\begin{table}[h!]
\centering
\begin{tabular}{|c|c|c|c|c|c|}
\hline
\textbf{Station} & \textbf{pre\_run} & \textbf{Execution time} & \textbf{post\_run} & \textbf{Acc}& \textbf{Sens}\\ \hline
1 & 0 / 0MB / 10s & 11h 26m 29s & 8 / 625 MB / 1min 2s & 0.737 & 0.737\\ \hline
2 & 8 / 625 MB / 59s & 12h 47m 24s & 13 / 625 MB / 1m 5s & 0.829 &  0.784\\ \hline
3 & 13 / 625 MB / 57s & 12h 55m 22s & 3 / 654.64MB / 1m 1s & 0.681 & 0.69\\ \hline
\textbf{Total} & time 2m 6s & 1d 13h 9m 8s & time 3m 25s  & 0.68 & 0.69 \\ \hline
\textbf{Central} & 0 / 0MB / 8.817s & 3d 23h 46m 25s & 3 / 654.64MB / 1m 1s & 0.684 & 0.625 \\ \hline
\end{tabular}

\caption{Execution time and performance of ISIC showcase model at different stations. \textit{pre\_run} and \textit{post\_run} protocols are security protocol steps. Number of files / file size / execution time is reported at each station in the protocol columns. Weighted accuracy (Acc) and weighted sensitivity (Sens) is averaged over all classes and reported from the last epoch at each station.} 
\label{exp2:time}
\end{table}

\textbf{Performance comparison of centrallized and distributed trains}
\begin{figure}[h!]

  \begin{minipage}[t]{0.49\textwidth}
	\centering
    \includegraphics[width=\textwidth]{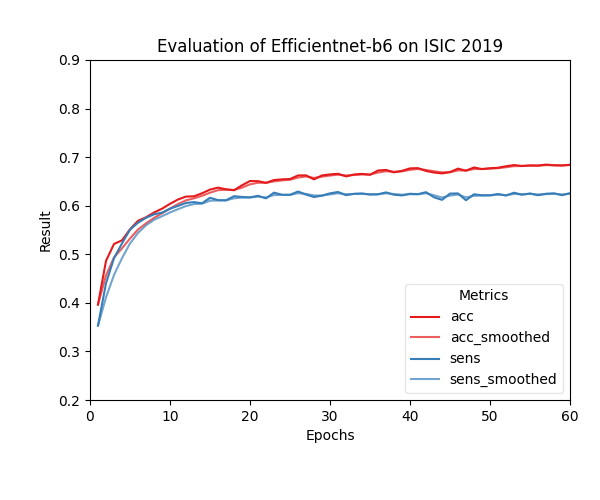}
	\caption{Averaged performance over 5 folds of a centrally trained model with all data available at one station trained for 60 epochs. Red lines indicate accuracy and blue lines sensitivity. Exponential moving average is implemented in tensorboard \cite{tensorflow2015-whitepaper} for the smoothed lines.}
	\label{exp:train2_central}
  \end{minipage}
    \hfill
  \begin{minipage}[t]{0.49\textwidth}
  	\centering
    \includegraphics[width=\textwidth]{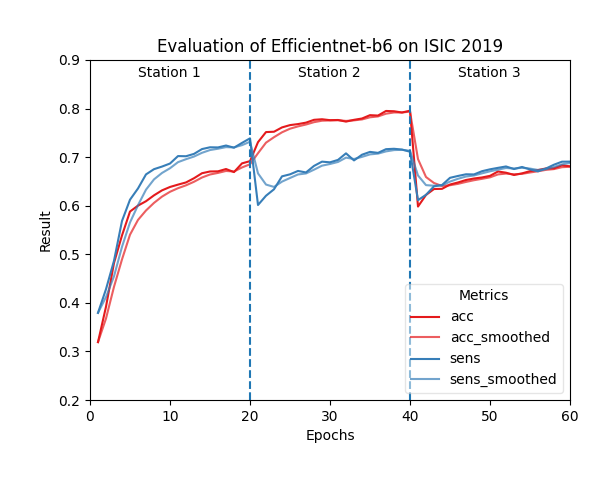}
	\caption{Averaged performance over 5 folds of a decentrally trained model with all classes \textbf{equally distributed} over three stations. Training was performed for \textbf{20 epochs} at each station. The same color schema and metrics are used as in figure \ref{exp:train2_central}.}
	\label{exp:train2_decentral}
  \end{minipage}
\end{figure}

\section{Conclusion}
Our proposed architecture of the PHT enables complex analyses in a secure way on patient data distributed across several hospitals.
At no time, patient data is leaving the control of the hospital.
Our security concept guarantees that only stations participating in the analysis and the creator of the train can decrypt the results.
Furthermore, any manipulation of queries and algorithms during execution or transit will be detected.
By relying on container technologies, this architecture can easily be extended to the rapidly changing requirements and particular needs of researchers.
This means complex analysis pipelines can be executed without the need to install  any additional software at hospitals.
Our software is deployment ready and freely available under permissive open-source licenses.
The two showcases demonstrate the flexibility of our PHT-meDIC solution by showing that, without additional configuration or installation efforts, it can be used for completely different analyses, a non-trivial task for other existing solutions.
Any kind of online analysis, varying from basic statistics to privacy preserving methods, can be executed using the same architecture.

\section{Discussion and Outlook of PHT}
For now our architecture is executing trains in sequential order.
Secure federated execution will be part of the next releases, following an extension of the security protocol and focusing on interoperability.
Working towards making trains a truly independent unit of processing allowing the execution regardless of the installed station software and will increase the acceptance and use of the PHT.
Additionally, DataSHIELD and vantage6 developers started collaboration in order to leverage the best of these two solutions\cite{van-ds-ps}.
Moving forward we will extend our architecture from the online setting to a federated design, improving our architecture's deep learning capabilities while maintaining our principles of security by design. Our aggregation protocol will guarantee data and model privacy under a fully adversarial threat model, closing the gap between SMPC and PHT. Furthermore, the problem of crucial forgetting shown in the second demonstration train will be solved using a federated design.
Today the PHT-meDIC supports the Paillier cryptosystem, where multiple hospitals can perform privacy-preserving computations.
Researchers can request to extend the train library with additional privacy methods (e.g., k-anonymity, differential privacy).

\section{Author Contributions}
M.H. contributed to software development, carried out experiments, and wrote the paper. M.G., P.P., F.K., F.B., and D.H. contributed to software development. L.Z. and M.S.contributed to the software. M.A. contributed to security aspects of the system. C.M. contributed to experiments. S.B., N.P. and O.K designed the study and wrote the paper.
All authors discussed and commented on the manuscript.

\subsection{Conflict of interest statement}
None declared.

\subsection{Code, Data, and Materials Availability}
Code for the PHT and experiments: \href{https://github.com/PHT-Medic}{https://github.com/PHT-Medic}.
Imaging data: \href{https://challenge2019.isic-archive.com/data.html}{isic 2019 challange data}.
1000 genomes data: \href{https://www.ncbi.nlm.nih.gov/variation/tools/1000genomes/}{ncbi.nlm.nih.gov}

\section{Funding}
Funding from German Federal Ministry of Education and Research (BMBF) within the ‘Medical Informatics Initiative’ [DIFUTURE, 01ZZ1804D; POLAR, 01ZZ1910K; CORD, 01ZZ1911O] and BMG [ZMVI1, 2520DAT94F] are gratefully acknowledged.


\bibliography{article}   
\bibliographystyle{splncs04}


\section{Supplementary Material}
\begin{enumerate}
    \item Used abbreviations in this paper. In cursive notations from the security protocol.
        \begin{table}[H]
        \centering
        \begin{tabular}{|l|l|}
        \hline
        \textbf{Term}     & \textbf{Description} \\ \hline
        $A$ & Algorithm files defined by the User \\ \hline
        AK & Actinic keratosis \\ \hline
        BCC & Basal cell carcinoma \\ \hline
        BKL & Benign keratosis (solar lentigo / seborrheic keratosis / lichen planus-like keratosis) \\ \hline
        $C_I$    & Container created of image $I$ \\ \hline
        $CR$ & Container Registry \\ \hline
        $D_{i}$ & Data (A,Q, and Model / Results) of party $i$ as cargo of the train\\ \hline
        DF & Dermatofibroma \\ \hline
        DNN & Deep Neural Network \\ \hline
        $DS_i$ & Digital Signature of party $i$ \\ \hline
        $\mathcal{E}_{D}$ & Encrypted value of data $D$\\ \hline
        GB & Gigabytes \\ \hline
        GPU & Graphical Processing Unit \\ \hline
        h & hours \\ \hline
        HLA & Human Leukocyte Antigen \\ \hline
        $I$ & Base image \\ \hline
        $ID_i$ & $ID$ of party $i$ \\ \hline
        $ID_U$ & Identifier of user $U$ \\ \hline
        $K$ & Random generated number of length $l$ as session key of the analysis \\ \hline
        KB & Kilobytes \\ \hline
        $N$ & Random generated number of length $l$ as session $ID$ of the analysis\\ \hline
        NV & Melanocytic nevus \\ \hline
        m & minutes \\ \hline
        MEL& Melanoma \\ \hline
        MHC& Major Histocompatibility Complex \\ \hline
        PDR& Private Docker Registry \\ \hline
        PHT& Personal Health Train \\ \hline
        $PK{_i}$ & Public key of the party $i$ \\ \hline
        $Q$ & Query operated on database defined by the User\\ \hline
        $R$ & Defined Route of the train defined by the User\\ \hline
        RE & Result extraction process\\ \hline
        $S$ & Station \\ \hline
        SCC & Squamous cell carcinoma \\ \hline
        $SK{_i}$ & Private key of the party $i$  \\ \hline
        SMPC & Secure Multi-Party Computation \\ \hline
        TB & Train building process \\ \hline
        $U$ & User  \\ \hline
        $UI$ & central User Interface do manage trains and submit algorithms\\ \hline
        $URI$ & Uniform Resource Identifier \\ \hline
        VASC & Vascular lesion \\ \hline
        vCPU & virtual Central Processing Unit \\ \hline
        \end{tabular}\label{supp:notations}
        \caption{Notations and abbreviations used in this paper.}
        \end{table}

    \item Detailed overview of execution: \textit{Architecture.png}
    \begin{figure}[h!]
  	\centering
    \includegraphics[width=\textwidth]{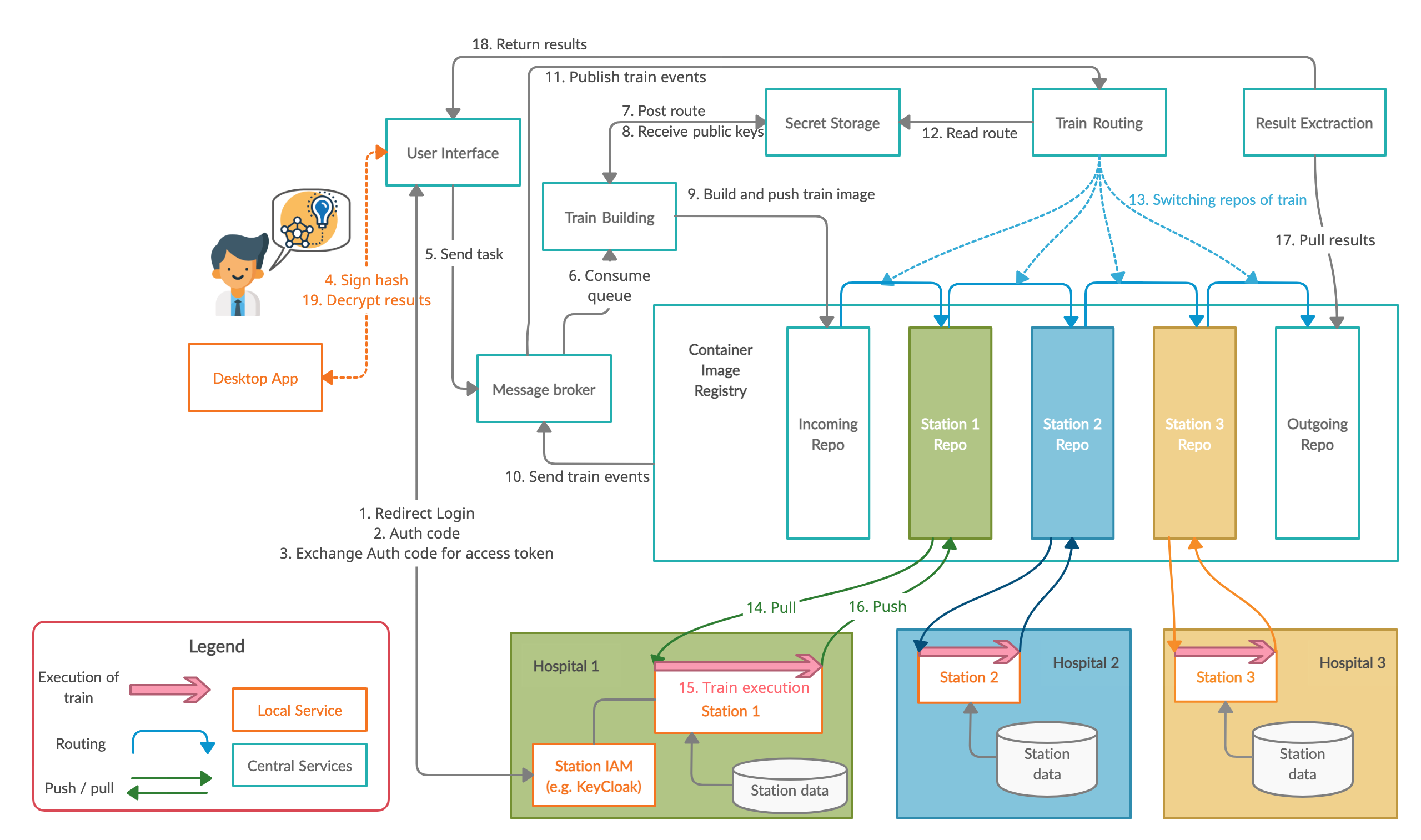}
	\caption{Extended overview of central service interactions to execute a train over three stations. Big pink arrows indicate the execution of trains locally at each hospital. Lines within the boxes indicate the update of results. Numbers indicate the order of processes.}
    \end{figure}
    
    \item \textbf{PHT security protocol}\label{supp:security}\\
    \textbf{Preliminaries of the security protocol}\\
    For the security concept the following definitions of asymmetric encryption function, symmetric encryption function and hash function are used:
    \begin{itemize}
        \item $\mathsf{ENCP}()$ is an asymmetric function that encrypts a message $M$ using a RSA public key $P$. 
    
        \item $\mathsf{DECP}()$ is an asymmetric function that decrypts a ciphertext $C$ encrypted with a $PK$ using corresponding RSA private key $SK$. 
        
        \item $\mathsf{ENCS}()$ is a function that encrypts a message $M$ under the given symmetric key $K$. 
        
        \item $\mathsf{DECS}()$ is a function that decrypts a ciphertext $C$ under the given symmetric key $K$. 
        
        \item $\mathsf{HASH}()$ is a cryptographic hash function which takes a message of an arbitrary length and produces a fixed length output. 
        For a given output $y_i$, it is computationally infeasible to find an input $x_i$ satisfying $\mathsf{HASH}(x_i)=y_i$. 
        
        \item $\mathsf{SIGN}()$ is a function that cryptographically signs the message $M$ using a private key $SK$ of a sender, the output can not be replicated without the private key and the content of $M$ verified 
        
        \item $\mathsf{VRFY}()$ is a function that takes three inputs: a message $M$, a ciphertext $C$ which is the ecryption of $HASH(M)$ with $SK$, and a public key $PK$ of a sender. It returns $\mathsf{HASH}(M) \stackrel{?}{=} \mathsf{DECP}(C,PK)$. 
        
    \end{itemize}
    
    \textbf{Security protocol assumptions}\label{prop_sec}
    \begin{itemize}
    \item The train registry can not change the predefined order of the stations participating in the analysis. It can not access the unencrypted FHIR search query. It can not access the intermediary results and the final result of the analysis.
    \item Data stations have to manually review and accept the algorithm and query. We assume that they are malicious parties.  An adversary compromising only one data station can access the intermediary results and the final result. The adversary can not access the data of other stations. 
    \item The user is malicious. If she or he collaborates with the private registry, the registry can access the query or intermediate results.
    \end{itemize}
    The notations used in the following section is provided in Table \ref{supp:notations} in the  supplementary material.
    \includecomment{Single point of failure of central service - will be solved in PHT 2.0 and cannot be done trivially within PHT 1.0}
    \\
    \textbf{Train proposal and submission}\\
    The following steps are required to submit secured trains:
    \begin{enumerate}
        \item A user $U$ uses his private credentials from his local governed identity management system to log into the central $UI$ to create a train. $U$ locally creates initially a private $SK_{U}$ and public $PK_{U}$ key pair using the Desktop App. He uploads his $PK_{U}$ to the central UI. $TB$ has a private $SK_{TB}$ and public $PK_{TB}$ key pair to sign users signatures. 
        \item The central service will upload the public key to vault.
        \item $U$ asks all stations to accept or reject a train proposal. All stations have unique pseudo identifiers (PIDs) within the PHT-meDIC, only the $UI$ knowing the matching real identifier.

        \item Based on the train proposal approval, $U$ specifies the included stations for his train. 
        \item $U$ provides algorithm $A$, query $Q$ and included stations $S_i$ in the $UI$.  
        $UI$ matches the station's ID to the corresponding $PIDs$ and sends the $TB$ the $PIDs$, uploaded $A$ and the public key of the user $PK_U$.
        \item $TB$ pulls one of the trusted base images $I$ of the train from the public read-only accessible section of the container registry $CR$.
        \item $TB$ generates the route $R=(PID_{S_1},PID_{S_{2}},...,PID_{S_{j}})$ and posts $R$ to vault and receives the public keys of the stations $PK_S$.
        \item $TB$ picks a random number $N \in \{0,1\}^l$ that is the session id of the analysis. 
        \item $TB$ picks a random number $K_U \in \{0,1\}^l$ that is the session key of the analysis.
        \item $TB$ calculates the hash (equation \ref{eq:hash}) of $(ID_U||A||Q||R||N)$ where $A$ is the algorithm, $Q$ is the query and $ID_U$ is the identifier of $U$. 
        
        \begin{equation}\label{eq:hash}
        H_U = \mathsf{HASH}(ID_U||A||Q||R||N)
        \end{equation}
        
        \item $TB$ provides $U$ with $H_U$
        \item $U$ signs $H_U$  (equation \ref{eq:sign}) with its private key $SK_U$ using the Desktop App and returns the signature $\mathcal{SIG}_{H_U}$ to the $U$, which will forward it to $TB$.
        \begin{equation}\label{eq:sign}
        \mathcal{SIG}_{H_U} = \mathsf{SIGN}(H_U,SK_{U})
        \end{equation}
        \item $TB$ encrypts (equation \ref{eq:enc} and \ref{eq:enc_q}) $Q$ and $A$ with the symmetric session key $K_U$. 
        
        \begin{equation}\label{eq:enc}
        \mathcal{E}_Q = \mathsf{ENCS}(Q,K_U)
        \end{equation}
        \begin{equation}\label{eq:enc_q}
        \mathcal{E}_A = \mathsf{ENCS}(A,K_U)
        \end{equation}
        \item $TB$ signs $\mathcal{SIG}_{H_U}$ (equation \ref{eq:sign}) with its private key $SK_{TB}$.
        \begin{equation}\label{eq:sign}
        \mathcal{SIG}_{TB} = \mathsf{SIGN}(\mathcal{SIG}_{H_U},SK_{TB})
        \end{equation}
        \item $TB$ encrypts (equation \ref{eq:enc_2}) $K_U$ with the public key of the all stations in $R$.
        
        \begin{equation}\label{eq:enc_2}
        \mathcal{E}_{K_U} = \{\mathsf{ENCP}(K_U,PK_{S_1}), \mathsf{ENCP}(K_U,PK_{S_{2}}),...,\mathsf{ENCP}(K_U,PK_{S_{j}})\}
        \end{equation}
        
        \item $TB$ builds an image  $I_0$ and writes
        $(ID_U,\mathcal{E}_A,\mathcal{E}_Q,R,N,\mathcal{E}_K,\mathcal{SIG}_H,\mathcal{SIG}_{TB})$ within.
        $I_0$ is named by a universally unique identifier (UUID) generated by the $UI$ logic.
        \item $TB$ pushes $I_{0}$ to the incoming repository of the $CR$.
    \end{enumerate}
    
    \textbf{Train execution}\\ 
    The previous valid built train can be executed by stations with these steps:
    \begin{enumerate}
        \item The train routing $TR$ process reads the route from vault and moves the image $I_{0}$ to the corresponding station repository.
        \item The data station $S_i$ pulls $I_{i}$ from it`s own station repository. 
        
        \item $S_i$ checks whether $I_{i}$ was built using one of the trusted base images.
        \item $S_i$ starts the $pre\_run$ protocol (algorithm \ref{train_pre_run}) by running $I_{i}$ 
       
        \begin{algorithm}[H]
        \caption{\textit{pre\_ run()} protocol}\label{train_pre_run}
        \begin{algorithmic}[1]
        \begin{small}
        \Procedure{pre\_run($ID_{S_i}$, $SK_{S_i}$)}{}
        \State $Q = \mathsf{DECS}(\mathcal{E}_Q,K_U)$
        \State $A = \mathsf{DECS}(\mathcal{E}_A,K_U)$
        \State $\text{valid} = \mathsf{VRFY}(ID_U||A||Q||N, \mathcal{SIG}_{H_U}, PK_U,PK_{TB})$
        
        \If {\textbf{not} \textit{valid}} abort train \EndIf
        \If {\textit{first station}}
        \State \textbf{run} train
        
        \Else
        \For {$j=i-1,i-2,\ldots,1$}
        \State $\text{valid} = \mathsf{VRFY}(DS_{j}[0], DS_{j}[1], PK_{j})$
        \If {\textbf{not} \textit{valid}} abort train \EndIf
        \EndFor
        \State $K_{i-1} = \mathsf{DECP}(\mathcal{E}_{K_{i-1}}[i],SK_{S_i})$ 
        \State $D_{i-1} = \mathsf{DECS}(\mathcal{E}_{D_{i-1}},K_{i-1})$
        \State $\text{valid} = \mathsf{VRFY}(\mathcal{D}_{i-1}||N, \mathcal{SIG}_{H_{D_{i-1}}}, PK_{S_{i-1}})$
        \If {\textbf{not} \textit{valid}} abort train \EndIf
        
        \State \textbf{run} train
        \EndIf
        \EndProcedure
        \end{small}
        \end{algorithmic}
        \end{algorithm}
        
        \item $S_I$ rebases decrypted results $D_{i-1}$ and executes train container $C_{I_{1}}$
        \item $S_I$ obtains results $D_i$ and runs the $post\_run$ protocol (algorithm \ref{train_post_run})

        \begin{algorithm}[H]
        \begin{small}
        \caption{\textit{post\_ run()} protocol}\label{train_post_run}
        \begin{algorithmic}[1]
        \Procedure{post\_run($D_{i}$, $ID_{Si}$, $SK_{S_i}$)}{}
        
        \State $H_{D_{i}} = \mathsf{HASH}(\mathcal{D}_i||N)$
        \State $\mathcal{SIG}_{H_{D_{i}}} = \mathsf{SIGN}(H_{D_{i}}, SK_{S_i})$
        \State $K_i = \{0,1\}^l$
        \State $\mathcal{E}_{\mathcal{D}_i} = \mathsf{ENCS}(\mathcal{D}_i,K_i)$
        \State $\mathcal{E}_{K_{i}} = \{\}$
        \For{$j = 1,2,\dots,n$}
        \State $\mathcal{E}_{K_{i}}[j] = \mathsf{ENCP}(K_i,PK_{S_j})$
        \EndFor
        \State $\mathcal{E}_{K_{i}}[n+1] = \mathsf{ENCP}(K_i,PK_{U})$
        \If {\textit{first\_station}}
        \State $DS_{i}= \mathsf{SIGN}(\mathsf{HASH}($N$), PK_{S_{i}})$
        \Else
        \State $DS_{i} = DS_{i-1}||\mathsf{SIGN}(\mathsf{HASH}(DS_{i-1}),PK_{S_{i}})$
        \EndIf
        \State return $\mathcal{E}_{\mathcal{D}_i}, \mathcal{E}_{K_i}, DS_{i}, \mathcal{SIG}_{H_{D_{i}}}$
        \EndProcedure
        \end{algorithmic}
        \end{small}
        \end{algorithm}
        
        \item $S_i$ writes $\mathcal{E}_{\mathcal{D}_i},\mathcal{E}_{K_i},DS_{i}$ to $C_{I_{0}}$. $S_i$ commits $C_{I_{0}}$ and creates an image $I_i$. $I_i$ is pushed to the stations own repository. 
        \item $TR$ moves the the successfully executed train of $S_i$ to the next stations repository according to the route.
         
    \end{enumerate}
    
    \item Table of subject in experiments 1: \textit{Experiment\_1\_Subjects.csv}\label{sup:subjects} 
    \begin{small}
    \begin{longtable}[H]{|p{0.2\linewidth}|p{0.2\linewidth}|p{0.2\linewidth}|}
\hline
\textbf{Subject} & \textbf{Probe} & \textbf{Station} \\ \hline
NA06985 & SRR709972 & 1\\
NA06994 & SRR070528 & 1\\
NA06994 & SRR070819 & 1\\
NA07000 & SRR766039 & 1\\
NA07048 & SRR099452 & 1\\
NA07056 & SRR764718 & 1\\
NA07357 & SRR764689 & 1\\
NA07357 & SRR764690 & 1\\
NA10847 & SRR070531 & 1\\
NA10847 & SRR070823 & 1\\
NA10851 & SRR766044 & 1\\
NA11829 & SRR710128 & 1\\
NA11831 & SRR709975 & 1\\
NA11832 & SRR766003 & 1\\
NA11840 & SRR070532 & 1\\
NA11840 & SRR070809 & 1\\
NA11881 & SRR766021 & 1\\
NA11992 & SRR701474 & 1\\
NA11994 & SRR701475 & 1\\
NA11995 & SRR766010 & 1\\
NA12003 & SRR766061 & 1\\
NA12004 & SRR766059 & 1\\
NA12005 & SRR718067 & 1\\
NA12006 & SRR716422 & 1\\
NA12043 & SRR716423 & 1\\
NA12043 & SRR716424 & 1\\
NA12044 & SRR766060 & 1\\
NA12144 & SRR766058 & 1\\
NA12154 & SRR702067 & 1\\
NA12155 & SRR702068 & 1\\
NA12156 & SRR764691 & 1\\
NA12234 & SRR716435 & 1\\
NA12249 & SRR070798 & 1\\
NA12716 & SRR081269 & 1\\
NA12717 & SRR071172 & 1\\
NA12750 & SRR081238 & 1\\
NA12750 & SRR794547 & 1\\
NA12750 & SRR794550 & 1\\
NA12751 & SRR071136 & 1\\
NA12751 & SRR071139 & 1\\
NA12760 & SRR081223 & 1\\
NA12760 & SRR081251 & 1\\
NA12761 & SRR077753 & 1\\
NA12761 & SRR081267 & 1\\
NA12762 & SRR718076 & 1\\
NA12763 & SRR077752 & 1\\
NA12763 & SRR081230 & 1\\
NA12812 & SRR715913 & 1\\
NA12813 & SRR718077 & 1\\
NA12813 & SRR718078 & 1\\
NA12814 & SRR715914 & 1\\
NA12872 & SRR716647 & 1\\
NA12874 & SRR764692 & 1\\
NA12878 & SRR098401 & 1\\
NA12891 & SRR098359 & 1\\
NA12892 & ERR034529 & 1\\
NA18501 & SRR100022 & 1\\
NA18502 & SRR764722 & 1\\
NA18502 & SRR764723 & 1\\
NA18504 & SRR100028 & 1\\
NA18505 & SRR716648 & 1\\
NA18505 & SRR716649 & 1\\
NA18507 & SRR764745 & 1\\
NA18507 & SRR764746 & 1\\
NA18508 & SRR716637 & 1\\
NA18508 & SRR716638 & 1\\
NA18516 & SRR100026 & 1\\
NA18517 & ERR034551 & 1\\
NA18522 & SRR107025 & 1\\
NA18523 & ERR034552 & 1\\
NA18526 & ERR031854 & 1\\
NA18532 & ERR031956 & 1\\
NA18537 & ERR032033 & 1\\
NA18537 & ERR032034 & 1\\
NA18542 & ERR031855 & 1\\
NA18545 & ERR031856 & 1\\
NA18547 & ERR031957 & 1\\
NA18550 & ERR031958 & 1\\
NA18552 & ERR031959 & 1\\
NA18558 & ERR031960 & 1\\
NA18561 & ERR031858 & 1\\
NA18562 & ERR031859 & 1\\
NA18563 & ERR031860 & 1\\
NA18566 & ERR031862 & 1\\
NA18571 & ERR031868 & 1\\
NA18572 & ERR031869 & 2\\
NA18573 & ERR031870 & 2\\
NA18576 & ERR031871 & 2\\
NA18577 & ERR032035 & 2\\
NA18577 & ERR032036 & 2\\
NA18579 & ERR032037 & 2\\
NA18579 & ERR032038 & 2\\
NA18582 & ERR031961 & 2\\
NA18592 & ERR031962 & 2\\
NA18593 & ERR034531 & 2\\
NA18603 & ERR031872 & 2\\
NA18608 & ERR031874 & 2\\
NA18609 & ERR031875 & 2\\
NA18620 & ERR031877 & 2\\
NA18621 & ERR034595 & 2\\
NA18622 & ERR032027 & 2\\
NA18622 & ERR032028 & 2\\
NA18623 & ERR032008 & 2\\
NA18624 & ERR031928 & 2\\
NA18632 & ERR031929 & 2\\
NA18633 & ERR031878 & 2\\
NA18635 & ERR031879 & 2\\
NA18636 & ERR031930 & 2\\
NA18853 & SRR100011 & 2\\
NA18858 & ERR034553 & 2\\
NA18861 & ERR034554 & 2\\
NA18870 & SRR100031 & 2\\
NA18871 & SRR100029 & 2\\
NA18912 & SRR111960 & 2\\
NA18940 & ERR034596 & 2\\
NA18942 & ERR034597 & 2\\
NA18943 & ERR034598 & 2\\
NA18944 & ERR034599 & 2\\
NA18947 & ERR034601 & 2\\
NA18948 & ERR034602 & 2\\
NA18949 & ERR034603 & 2\\
NA18951 & ERR034604 & 2\\
NA18952 & ERR034605 & 2\\
NA18953 & SRR099546 & 2\\
NA18959 & SRR099545 & 2\\
NA18960 & SRR099533 & 2\\
NA18961 & SRR099544 & 2\\
NA18966 & SRR071175 & 2\\
NA18967 & SRR071192 & 2\\
NA18967 & SRR071196 & 2\\
NA18968 & SRR077480 & 2\\
NA18968 & SRR081231 & 2\\
NA18969 & SRR081266 & 2\\
NA18969 & SRR081273 & 2\\
NA18970 & SRR071116 & 2\\
NA18970 & SRR071127 & 2\\
NA18971 & SRR077447 & 2\\
NA18972 & SRR077490 & 2\\
NA18972 & SRR081255 & 2\\
NA18973 & SRR077861 & 2\\
NA18973 & SRR078846 & 2\\
NA18974 & SRR077456 & 2\\
NA18974 & SRR081248 & 2\\
NA18975 & SRR078849 & 2\\
NA18976 & SRR077451 & 2\\
NA18976 & SRR077757 & 2\\
NA18978 & SRR716650 & 2\\
NA18981 & SRR077477 & 2\\
NA18981 & SRR077751 & 2\\
NA18987 & SRR077491 & 2\\
NA18990 & SRR077454 & 2\\
NA18990 & SRR077486 & 2\\
NA18991 & SRR077450 & 2\\
NA18991 & SRR077855 & 2\\
NA18992 & SRR716428 & 2\\
NA18994 & SRR716431 & 2\\
NA18995 & SRR764775 & 2\\
NA18997 & SRR702078 & 2\\
NA18998 & SRR766013 & 2\\
NA18999 & SRR112297 & 2\\
NA19000 & SRR099528 & 2\\
NA19003 & SRR099532 & 2\\
NA19005 & SRR715906 & 2\\
NA19007 & SRR099549 & 2\\
NA19012 & SRR112294 & 2\\
NA19092 & SRR100012 & 2\\
NA19093 & SRR100033 & 2\\
NA19098 & SRR077460 & 2\\
NA19099 & SRR748771 & 2\\
NA19099 & SRR748772 & 2\\
NA19102 & SRR100034 & 2\\
NA19116 & SRR100021 & 2\\
NA19119 & SRR077471 & 2\\
NA19119 & SRR081271 & 2\\
NA19130 & SRR107026 & 2\\
NA19131 & SRR070494 & 2\\
NA19131 & SRR070783 & 2\\
NA19137 & SRR081226 & 2\\
NA19137 & SRR081237 & 2\\
NA19137 & SRR792542 & 2\\
NA19137 & SRR792560 & 3\\
NA19138 & SRR070472 & 3\\
NA19138 & SRR070776 & 3\\
NA19141 & SRR077433 & 3\\
NA19141 & SRR077464 & 3\\
NA19143 & SRR077445 & 3\\
NA19143 & SRR081272 & 3\\
NA19144 & SRR077392 & 3\\
NA19152 & SRR071135 & 3\\
NA19152 & SRR071167 & 3\\
NA19153 & SRR070660 & 3\\
NA19153 & SRR070846 & 3\\
NA19159 & SRR070478 & 3\\
NA19159 & SRR070786 & 3\\
NA19171 & SRR077492 & 3\\
NA19171 & SRR077493 & 3\\
NA19172 & SRR111962 & 3\\
NA19200 & SRR077432 & 3\\
NA19200 & SRR078847 & 3\\
NA19201 & SRR077439 & 3\\
NA19201 & SRR077462 & 3\\
NA19204 & SRR077857 & 3\\
NA19204 & SRR081263 & 3\\
NA19206 & SRR070491 & 3\\
NA19206 & SRR070781 & 3\\
NA19207 & SRR081254 & 3\\
NA19207 & SRR081256 & 3\\
NA19209 & SRR077489 & 3\\
NA19209 & SRR077859 & 3\\
NA19210 & SRR078845 & 3\\
NA19210 & SRR081222 & 3\\
NA19238 & SRR071173 & 3\\
NA19238 & SRR071195 & 3\\
NA19238 & SRR792121 & 3\\
NA19238 & SRR792165 & 3\\
NA19239 & SRR792097 & 3\\
NA19239 & SRR792159 & 3\\
NA19240 & SRR792091 & 3\\
NA19240 & SRR792767 & 3\\\hline
\caption{Included subjects, probes and associated station for demonstration of nf-core pipeline with experiment \ref{exp:nf-core}}
\label{sup:subjects}
\end{longtable}
    \end{small}
    
\end{enumerate}

\end{spacing}
\end{document}